# A Matching Algorithm based on Image Attribute Transfer and Local Features for Underwater Acoustic and Optical Images

Xiaoteng Zhou, Changli Yu, Xin Yuan, Citong Luo

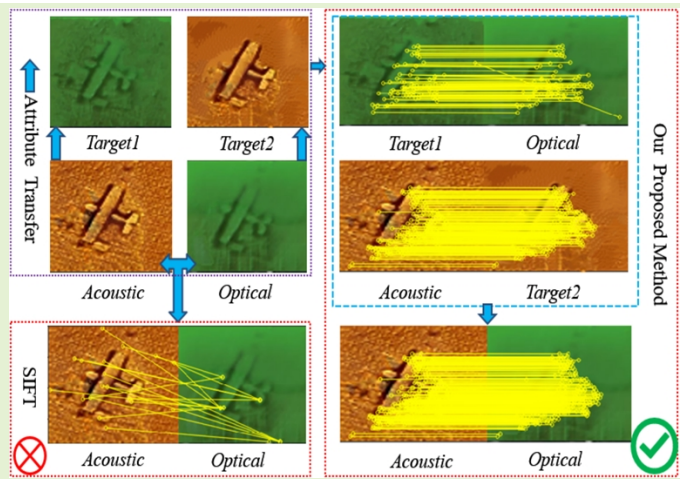

*Abstract*—In the field of underwater vision research, image matching between the sonar sensors and optical cameras has always been a challenging problem. Due to the difference in the imaging mechanism between them, which are the gray value, texture, contrast, etc. of the acoustic images and the optical images are also variant in local locations, which makes the traditional matching method based on the optical image invalid. Coupled with the difficulties and high costs of underwater data acquisition, it further affects the research process of acousto-optic data fusion technology. In order to maximize the use of underwater sensor data and promote the development of multi-sensor information fusion (MSIF), this study applies the image attribute transfer method based on deep learning approach to solve the problem of acousto-optic image matching, the core of which is to eliminate the imaging differences between them as much as possible. At the same time, the advanced local feature descriptor is introduced to solve the challenging acousto-optic matching problem. Experimental results show that our proposed method could preprocess acousto-optic images effectively and obtain accurate matching results. Additionally, the method is based on the combination of image depth semantic layer, and it could indirectly display the local feature matching relationship between original image pair, which provides a new solution to the underwater multi-sensor image matching problem.

*Index Terms*—Underwater vision, sensor data fusion, image matching, attribute transfer, image processing, deep learning

## I. INTRODUCTION

IN RECENT times, many countries gradually began to obtain renewable energy from the deep sea to meet the needs of humans and industries. With the gradual deepening development of marine resources, deep-sea exploration activities will become more frequent. As the primary detection vehicles, such as autonomous underwater vehicles (AUVs) and remotely operated vehicles (ROVs) are also equipped with more advanced acoustics and optical sensor. These sensors have played an outstanding role in seabed geomorphological mapping task, target recognition and classification, biological research, resource exploration, environmental monitoring and other fields [1-6]. Sonar is the most commonly used sensor in the field of deepwater exploration, which could collect images of sea targets at a relatively long distance and is not influenced by turbidity. However, it will encounter special cases in the imaging process, such as low signal-to-noise ratio (SNR) and low resolution, low feature repeatability, etc. [6]. In addition, there are problems such as blurred target edges, serious distortion and poor overall image quality in imaging. What's worse is when sonar detects the same area from different angles, these areas may have dramatic lighting changes [7]. Optical cameras could intuitively provide high-resolution and highly recognizable target images, but due to the scattering effects of light in seawater and the influences of the medium on its absorption, most of the optical images are blue-green and the imaging distance is severely limited. Even if the acoustic and optical sensors are in the same scene, they are diverse in terms of clarity, color, dynamic range, SNR, and structural similarity. In order to alleviate their respective limitations, fully combine their advantages. Acousto-optic data fusion technology has developed rapidly in recent years, this technique is active in marine observation and joint positioning applications, such as the use of optical image information, including color, contour, etc. to further explore the details of the target object in the acoustic image [8]. Committed to better solving the problem of acousto-optic image matching and serving the development of

Manuscript submitted July 12, 2021. This work was supported by the Chinese Shandong Provincial Key Research and Development Plan under Grant No. 2019GHZ011. *(Corresponding author: Changli Yu.)*

The authors in this manuscript are with the School of Ocean Engineering, Harbin Institute of Technology, Weihai 264200, China (e-mail: zhouxiaoteng@stu.hit.edu.cn; yuchangli@hitwh.edu.cn; xin.yuan@upm.es; luocitong@gmail.com).

underwater sensor data fusion technology, this letter proposed an image matching method based on image attribute transfer and learned descriptors. The method is divided into two steps:

Firstly, the method of image attribute transfer is used to eliminate the difference between acousto-optic images to the greatest extent to meet the needs of precise matching. It is worth mentioning that style transfer method has made great achievements in the field of computer vision [9-14]. In the above, some scholars have used the method of style transfer to deal with the differences of various modal imaging [15-19].

Secondly, the traditional matching ideas based on extracting regional local features such as points, lines and contours, such as in [20-22], the authors usually require hand-made features, which are easily affected by external changes, and most methods were originally developed for optical images, so that the performance on acoustic images is not satisfactory. Some researchers have tried using traditional descriptors [6,7,23,24]. However, usually these methods are suitable only for a specific sonar research and do not have a generalization. Additionally, the image processing is more cumbersome and requires rich professional experiences to intervene in the processing. It is difficult to form an effective and fixed processing method. Acousto-optic images, although the semantic structures of the image pairs are similar, there are huge differences in texture features and noise distribution, and in a complex underwater environment, the acousto-optic sensor will be accompanied by changes in angle and scale when imaging. The ensuing affine problem has become a difficulty of acousto-optic matching.

In the field of architectural and remote sensing image matching research, some scholars have used the learned descriptors to achieve exciting results, and the matching accuracy has surpassed the traditional local feature descriptors [25-30]. Therefore, we propose to introduce the learned descriptor HardNet [31,32] to solve the challenging underwater acousto-optic image matching task.

Additionally, in view of the difficulty and high cost of underwater acousto-optic data acquisition, in order to better meet the experimental requirements and the optimization design requirements of local feature descriptors, this paper uses the sonar style transfer idea proposed in [19] to carry out image simulation, the purpose is to generate realistic underwater acousto-optic images to expand training samples. This method satisfies the data synthesis requirements of multiple sonar types. Experimental results show that our proposed approach achieves fast and accurate feature point detection and local feature matching on the real and simulated underwater acousto-optic images with less human intervention. This method could be applied to a variety of scenarios and has a certain degree of robustness.

The main framework of the paper is as follows. Section II reviews classic and cutting-edge researches regarding underwater acousto-optic image matching. Section III presents the detailed methodology of our work. Section IV demonstrates the data for experiment. Section V verifies the algorithm and compares the evaluation results. Section VI makes the discussion and outlook. Conclusion is drawn in Section VII.

## II. RELATED WORK

The research inspiration for underwater acoustics and optical image matching is based on diverse sensor data fusion ideas. Image matching is a classic research topic. It is roughly divided into regional and local feature matching. Regional matching focuses on the correlation within a group of comparative image sets [8]. However, feature-based matching focuses on the details of the target object's points, lines, contours, etc.

In 2004, Fusiello and Murino [33] earlier proposed joint acousto-optic device used in underwater environment perception, and underwater scene modeling issues verify the effectiveness of the proposed ideas, Negahdaripour [34] introduced in more detail in the system calibration and 3D object reconstruction task, acousto-optic reveal sensor fusion technology advantages, and introduces a new method of photoacoustic stereo calibration. In the medical field, some scholars have also done a lot of research on acousto-optic information fusion technology [35].

High quality image acquisition is the foundation of acousto-optic data fusion technology research. For this, a lot of researches on improving and restoring underwater optical image have been done, such as in [36-38]. In [39,40], the restoration and feature extraction methods of sonar images have been studied.

In the field involving underwater sonar matching, Vandrish [7] compared the results of sidescan sonar image matching using scale-invariant feature transform (SIFT) and other traditional local feature descriptors, and concluded that SIFT performed best among the traditional matching methods. SIFT registration algorithm for two synthetic aperture sonar images was studied in [23], and two ideal sonar track image registration geometric models were proposed. Kim [24] proposed an idea of associating the detected key points with Harris corner detection in a general sonar image registration task. Hurtós [39] used features based on Fourier transform to match forward-looking sonar images and achieved satisfactory results, but this method was limited by the rotation and translation range. Toro [41] tried to use convolutional neural network (CNN) to learn the matching mapping of sonar images, and proposed an algorithm to generate matching pairs for training from the labeled target. This method could directly learn the matching function from the labeled data without any manual feature engineering. The final result shows that the accuracy of sonar image matching is higher after the feature processing of CNN, while the accuracy of the classical keypoint method is lower. Pham [42] used the guidance of block matching, a segmented sonar image with a self-organizing map for registration and Mosaic of sidescan sonar images. Yang [43] proposed an image matching method based on CNN algorithm, which aims to alleviate the contradiction between traditional image features and similarity learning. For example, when improving the background dynamics of sonar images, the low intensity and high matching accuracy are inconsistent. The model can be trained to learn the image texture features of sonar without any artificial design of feature descriptors.

Previous studies in sonar matching mainly focus on image processing of a single sensor, and sonar image matching studies mainly serve for image registration and Mosaic, for example [24,39]. In recent years, some scholars have begun to study how to combine other sensors (mainly optical) on the basis of sonar technology for matching research.

Liu [8] tried to design a region-based filter to deal with the problem of acousto-optic matching. This method could eliminate the impact of changes in the viewing angle and environmental changes during sensor imaging. At the same time, they proposed an iterative algorithm to enhance the image, and then increase the proportion of effective information in the graph, and then use morphological filtering for noise suppression. Finally, use Gaussian multiscale images to optimize the matching results in order to reduce scale errors. The experimental results show that this method could initially realize the macroscopic matching on the acousto-optical image area. However, this method requires the design of a variety of attached algorithms and filters, requires a lot of expert experience and manual intervention, and the overall process is relatively complicated, and it is impossible to match the details of the image. The lack of details of local feature matching has certain limitations in the research of improving the autonomy of deep-submersible vehicles.

Hyesu Jang [18] proposed a style transfer method based on CNN [9] and combined with traditional feature descriptors to match underwater acousto-optic images. The idea of this manuscript was also inspired by their creativity, but their method can only finish the matching in one style, such as generating a sonar image in an optical style, then finally matched in the optical style. Secondly, the parameter setting of their style transfer method is not clear, and there are many custom parameters. A lot of attempts in their idea are required during the experiment. The images before and after the transfer cannot establish a deep mapping relationship, besides in the process of image style transfer, the introduction of noise will easily destroy the original structure and details of the image.

This article proposed a new method for underwater acoustic and optical image matching. Based on the advanced image attribute transfer algorithm [10], we introduce a learned descriptor [31] with stronger expressiveness in the field of complex image matching. In the first step, we use the image attribute transfer method to eliminate the difference between acousto-optic images, and on the premise of ensuring the target content structure, we then look for areas with similar content structures in the two images, and gradually unify the colors, textures, and styles. The above preprocessing of the image paves the way for the subsequent introduction of learned descriptors for matching, and further strengthens the matching effect of acousto-optic images based on local features.

## III. METHOD

### A. Image Attribute Transfer

The visual attribute transfer method introduced in [10] has been widely used in image texture, color, content and style transfer scenes. The core of this method is to give full play to the superior performance of CNN's deep feature extraction in the image processing process. The deep pyramid features extracted by CNN are used to construct the semantic dense correspondence between image pair, and gradually adjust and optimize to achieve the attribute transfer of two images.

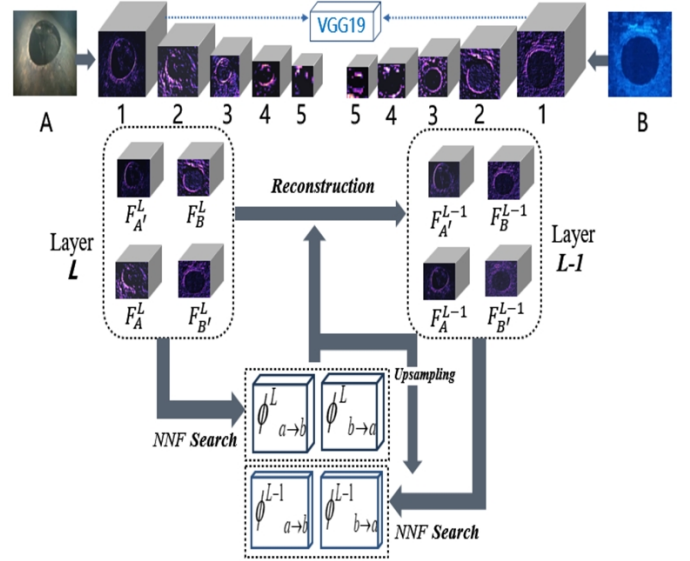

Fig. 1. The pipeline of the image attribute transfer.

As shown on the top of Fig. 1, we transmit sonar images and underwater optical images with relevant semantic information to the CNN network (VGG19 [44]) for the construction of feature pyramid. Then we select five groups of depth feature maps $\{F_A^L\}$ and $\{F_B^L\}$ ($L = 1\ldots 5$). From the first layer to the fifth layer of the VGG model, the image details are gradually lost, and almost only the high-level semantic information is left at the top layer. $F_A^L$ is a feature map which represents the response of image $A$ on the scale $L$ and $F_B^L$ has the similar definition in reverse direction, corresponding $F_{A'}^L$ indicates that, when reconstructed the image on scale $L$, the feature map has the content structure of image $A$ and style details of image $B$.

After establishing the mapping relationship at the coarsest layer ($L = 5$), the next step is to iterate from the high level to the low level. Its application in the field of acousto-optic integration is described as follows:

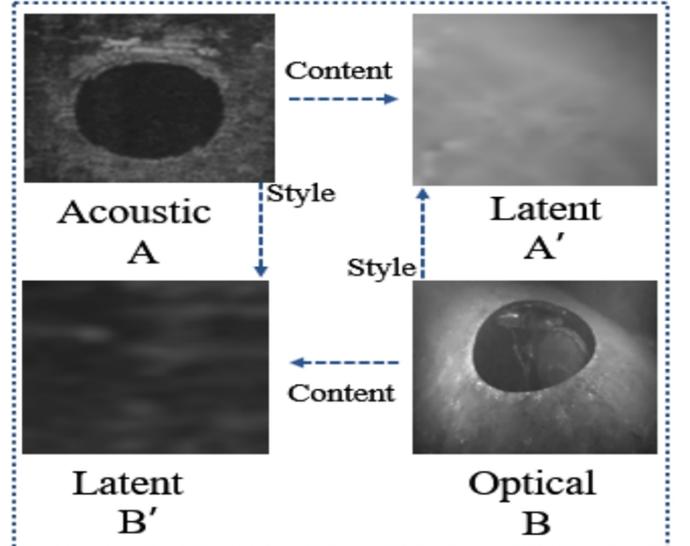

Fig. 2. The raw acoustic and optical image pair before attribute transfer.

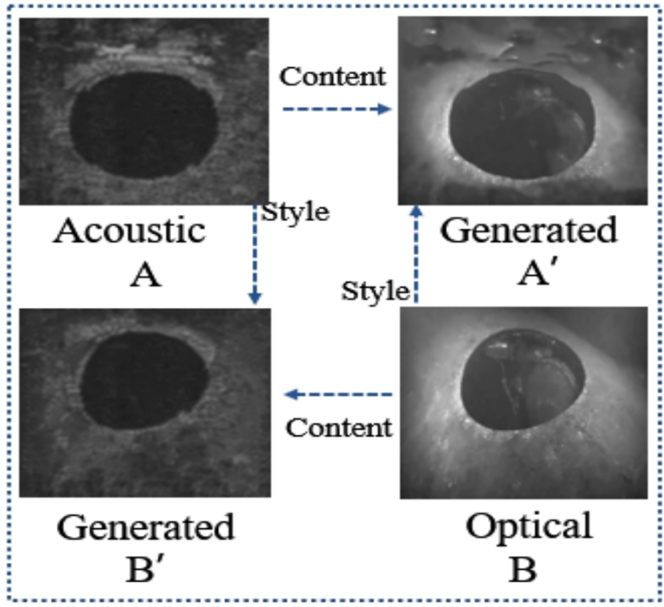

Fig. 3. The acoustic and optical image pair after attribute transfer.

Initially, the feature of *A'* and *B'* are unknown. We estimate them from coarse to fine, which requires good initialization at the coarsest layer ($L=5$). Assume $F_{A'}^L = F_A^L$ and $F_{B'}^L = F_B^L$ are satisfied in the coarsest layer. In this moment, it is equivalent to ignoring the detailed information of the image by default, and the semantic information is consistent. The process description is mainly divided into the following feature alignment and image reconstruction these two parts:

1) *Feature alignment*

Before image reconstruction, it is necessary to establish a position mapping between feature maps, that is, feature alignment. This method mainly adopts Nearest Neighbor Field (NNF) for feature alignment, in which the algorithm used to calculate the approximate NNF between two images is well applied in [45]. In this study, the similarity between patches is mainly considered, then use mapping functions $\phi_{a \to b}^L$ to represent a forward NNF we estimated and $\phi_{b \to a}^L$ represents a reverse NNF. In other words, $\phi_{a \to b}^L$ means to search the patch $p$ in $F_A^L$ and find another patch $q$ with the smallest distance from it in $F_B^L$. $\phi_{b \to a}^L$ is similarly defined in the reverse direction. The reference formula is as follows:

$$\phi_{a \to b}^L(p) = \arg_q \min \sum_{x \in N(p), y \in N(q)} \left( \varphi^2 + \phi^2 \right) \quad (1)$$

$$\varphi = \left\| \overline{F}_A^L(x) - \overline{F}_{B'}^L(y) \right\| \quad (2)$$

$$\phi = \left\| \overline{F}_{A'}^L(x) - \overline{F}_B^L(y) \right\| \quad (3)$$

where $N(p)$ is the patch around $p$, $F(x)$ is a vector in position $x$. $\overline{F}^L(x)$ represent normalized features in patch similarity metric, which are better for matching.

The equation (1) could easily obtain four variables by using the ideal assumption conditions at the coarsest layer, and then obtain the analogy mapping relationship between the acousto-optic images at the layer $L=5$. After that, this analogy mapping relation is used to derive between adjacent lower levels.

2) *Image reconstruction*

The intention of image reconstruction is that the ideal latent image $A'$ fully retains the content structure of the original sonar image $A$ while referring to the style details of the optical image $B$. The ideal latent image $B'$ has a similar explanation in the reverse direction, and the idea vividly illustrated in [9].

After the feature processing is completed, we start from the coarsest layer in the image layer to reverse the low-level layer. Until the first layer gets $F_{A'}^1$ and $F_{B'}^1$, which are the target images we want to generate. That is, based on the known mapping relationship of the layer $L$, the mapping relationship of the layer $L-1$ is gradually explored.

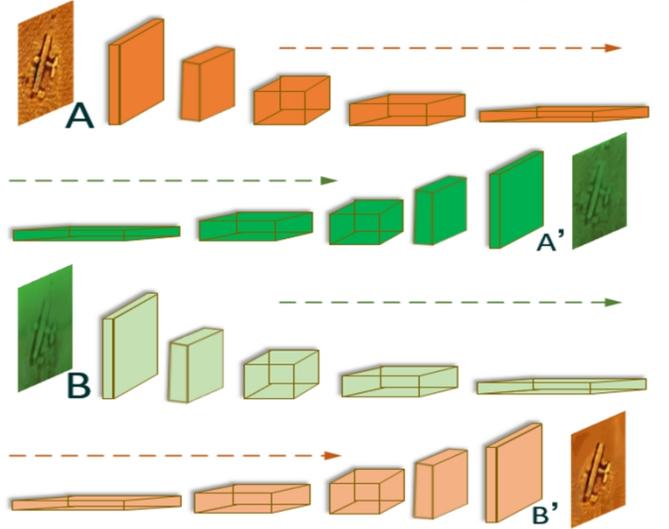

Fig. 4. Pipeline of image reconstruction.

Taking $A'$ generated by $A$ and $B$ as an example, based on the ideal assumptions of the coarsest layer ($F_{A'}^L = F_A^L$), deconvolution of $A'$ is carried out by using the feature map of layer $L$ of $A$ ($F_A^L$). To satisfy $A'$ maintaining the content structure of $A$ while incorporating the style details of $B$, the deconvolution result of $F_A^L$ needs to be adjusted before input to $F_{A'}^{L-1}$. In this adjustment, custom weight coefficient $W$ is introduced to control the similarity of $A'$ to $A$ and $B$, that is, $W$ weighs the proportion of content and style, and it is modified with the change of layer number $L$. Formula is as follows:

$$F_{A'}^{L-1} = F_A^{L-1} \circ W_A^{L-1} + R_B^{L-1} \circ \left(1 - W_A^{L-1}\right) \quad (4)$$

$$F_{B'}^{L-1} = F_B^{L-1} \circ W_B^{L-1} + R_A^{L-1} \circ \left(1 - W_B^{L-1}\right) \quad (5)$$

By using equation (4) for analysis, $R_B^{L-1}$ is obtained by deconvolution of $R_B^L$, and $R_B^L$ is obtained by deformation of $F_B^L$ in order to be close to $F_A^L$ in content structure. The deformation criterion is based on the double constraint $(\phi_{a \to b}, \phi_{b \to a})$ of $F_A^L$ and $F_B^L$, and matching is carried out according to the search result of NNF. Equation (5) is similarly defined in the reverse direction. The finally obtained $F_A^{L-1}$ and $F_B^{L-1}$ are combined with the feature of $A$ and $B$ at layer $L-1$, where $\circ$ is element-level multiplication.

After obtaining the above four custom variables of layer $L-1$: $F_A^{L-1}, F_{A'}^{L-1}, F_B^{L-1}$ and $F_{B'}^{L-1}$, the mapping relationship between images of this layer can be solved. Similarly, using this iterative method, we could finally get target images $A'$ and $B'$. It should be noted that we input two types of images and output two types of images. Using this method to establish the pixel location mapping relationship, the image $A'$ is based on the content of the sonar image and the style is referenced to the optical image, while the image $B'$ is on the opposite side.

### B. Learned Descriptor

In view of the self limitations of the underwater acoustic images and optical images proposed above, we tried to introduce the current very advanced learned descriptor HardNet in this research to solve underwater image matching task.

HardNet is inspired by the Lowe's matching criterion for SIFT. This method introduces a new loss for metric learing based on the CNN structure of L2-Net [46]. Experiments show that it has the same dimension as SIFT and demonstrates the most advanced performances in terms of wide baseline stereo vision, patch validation and instance retrieval benchmark.

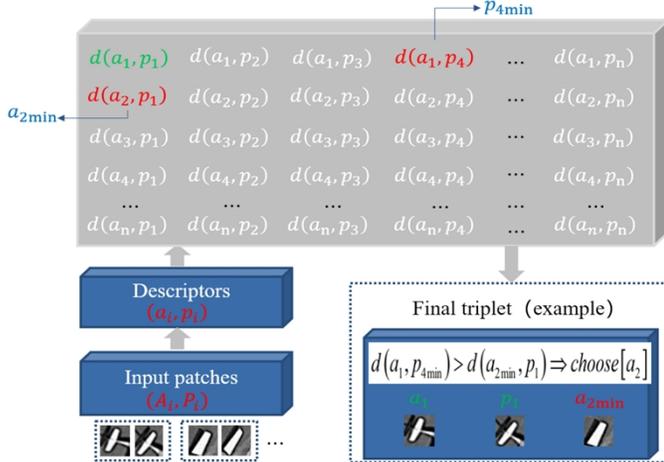

Fig. 5. Schematic diagram of the HardNet sampling procedure.

One training batch contains two batches and $n$ matching pairs, each of which is represented by $(a_i, p_i)$ respectively. The $L_2$ distance matrix is as follows:

$$D = c\text{dist}(a, p) \quad (6)$$

$$d(a_i, p_j) = \sqrt{2 - 2a_i p_j}, i = 1 \ldots n, j = 1 \ldots n \quad (7)$$

This method adopts the network structure of L2-Net and has been trained on Brown dataset [46]. Through several training experiments, the performance of HardNet in matching tasks has surpassed that of traditional local descriptors and other learned descriptors [31].

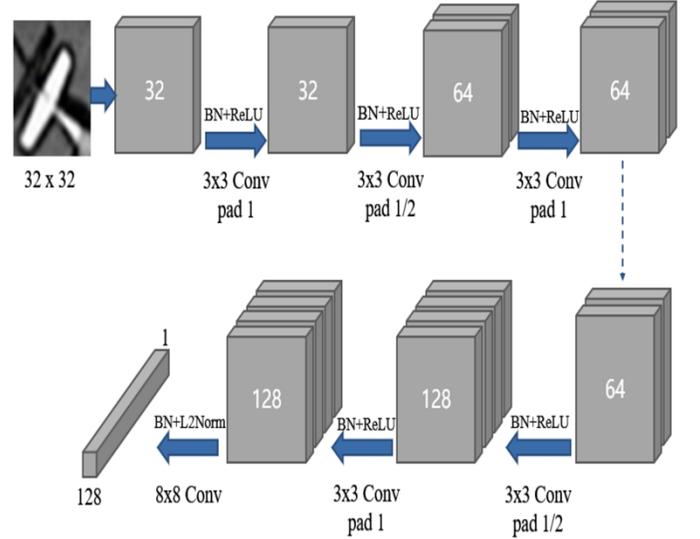

Fig. 6. The network architecture of HardNet.

The procedure of the underwater acoustic and optical image matching algorithm is expressed as follows.

---
**ALGORITHM 1** : Underwater acoustic and optical image matching algorithm (UAOM)
---
**Input**: Underwater sonar image *A* and optical image *B*
**Output**: Matching result of image *A* and image *B*
**Preprocessing**: Utilize VGG19 model to analyze *A* and *B*, get deep features $F_A^L$ and $F_B^L$, $F_{A'}^5 = F_A^5, F_{B'}^5 = F_B^5, L = 5$, and randomize mapping function $\phi^5_{a \to b}$ and $\phi^5_{b \to a}$

**For** $L \in [5,1]$ **do**
    **NNF search**: $\phi^5_{a \to b}$ and $\phi^5_{b \to a}$
    **If in the queue** $L > 1$ **then**
        **Begin reconstruction**:
        Warp $F_B^L$ with $\phi^L_{a \to b}$
        Deconvolve $F_B^L$ to $R_B^{L-1}$
        Weighted sum of $F_A^{L-1}$ and $R_B^{L-1}$ to $F_{A'}^{L-1}$
        Warp $F_A^L$ with $\phi^L_{b \to a}$
        Deconvolve $F_A^L$ to $R_A^{L-1}$
        Weighted sum of $F_B^{L-1}$ and $R_A^{L-1}$ to $F_{B'}^{L-1}$
        **NNF upsampling**:
        $\left(\phi^L_{a \to b} \to \phi^{L-1}_{a \to b}\right)$ and $\left(\phi^L_{b \to a}, \phi^{L-1}_{b \to a}\right)$
    **END**
**END**
Generate potential images *A'* and *B'*
**MATCHING:**
    Utilize the learned descriptor HardNet to match *A* and *B'*, *A'* and *B*
    Gain the mapping correspondences bettween *A* and *B*
**END**

## IV. EXPERIMENT

### A. Test Data Sets

In order to verify the effectiveness of our method, we select 6 groups of image pairs for testing experiments, and the underwater image information is shown in Fig. 7.

Image pair 1 and 2 are paver bricks and a hatch of one sunken plane, respectively, in which the acoustic images of them are captured by the ARIS Explorer 3000 which is a Dual Frequency Identification Sonar (DIDSON), all the original images are provided by SOUND METRICS [47].

Image pair 3 is a mine shape and image pair 4 is a letter string, their acoustic images both are obtained by DIDSON, which are derived from [48] and [18] respectively.

Image pair 5 is generated using the style transfer method proposed in [19], in which the style of the acoustic image is transferred DIDSON, and the optical image is the real image.

Image pair 6 is also generated using the method proposed in [19], in which the style of the acoustic image is transferred to the more common side scan sonar (SSS), and the optical style is transferred to the mode of underwater optics. In order to increase the contrast effect, the RGB mode is used to display.

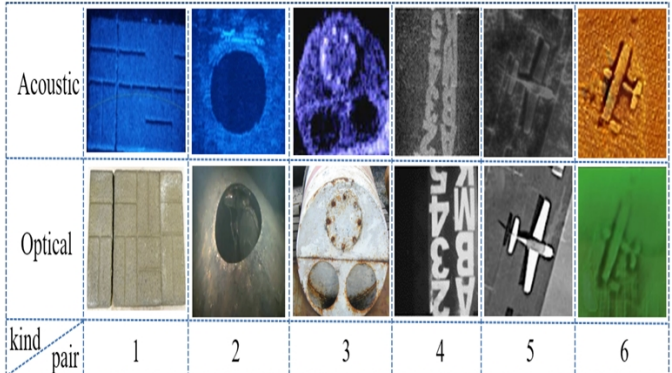

Fig. 7. The test samples set: (1) paver bricks, affine transformation (2) a hatch, affine transformation (3) a mine shape, affine transformation (4) a letter string, affine transformation (5) plane, DIDSON style (6) plane, SSS and underwater optical style.

### B. Experimental Control Groups Sets

In terms of the control experiment, we introduce the classical image local feature descriptors SIFT, SURF and BRSIK [49] as the comparison, which respectively have excellent performance in robustness and speed, especially for sonar and other images with large blur. At the same time, in order to make the performance of HardNet descriptor more expressive, we introduce HesAffNet [32] as the detector to complete the matching task together.

## V. RESULTS AND EVALUATION

### A. Evaluation Indexes Sets

We introduce the number of good matches (GM), average number of inliers (INL) per matched pair, matching accuracy (MA) and running time (RT) four indexes to evaluate match methods. GM measures the robustness, INL and MA measure the accuracy of the algorithm and RT measures the real-time performance of the algorithm. For each image pair, we take the indexes average of ten test results as the final evaluation results.

1) *GM*

We adopt the number of good matches in per image pair when the *ratio* is 0.8 to measure the adaptability and robustness of the method. The larger the number of GM obtained for each group of images, the better the performance of the matching method.

2) *INL*

We take the average number of inliers in per image pair when the *ratio* is 0.8 to reflect the accuracy of our proposed method, the higher the value, the better the performance.

3) *MA*

Introduce the matching accuracy to reflect the effective utilization of our algorithm, MA is numerically equal to the ratio of INL to GM. To a certain extent, MA could reflect the coordination between the detector and the descriptor.

4) *RT*

In underwater engineering operations, real-time operation is a fixed requirement, so we introduce RT as the time evaluation index to measures the matching time, so as to verify the complexity of our algorithm.

### B. Test Tools and Environment Details

All methods were implemented under the Windows 10 operating system using Python 3.7 with an Intel Core i7-9700 3.00GHz processor, 16GB of physical memory, and one NVIDIA GeForce RTX2070s graphics card. SIFT, SURF and BRISK are implemented based on openCV-Python tools [50]. In order to better display the matching results of local features of acousto-optic images, the grayscale display mode is adopted.

### C. Test and Evaluate Results

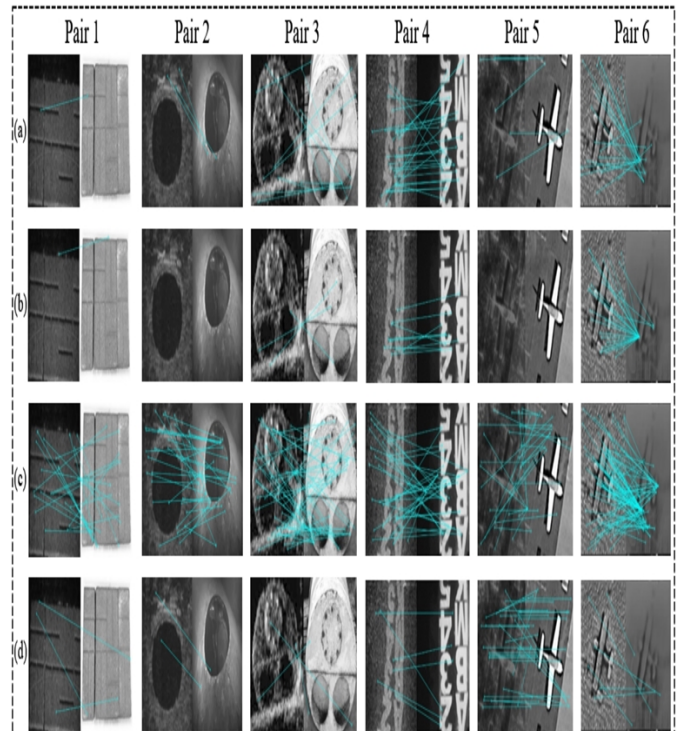

Fig. 8. This is a schematic diagram of the matching results of the raw image pairs: (a) Matching results using SIFT; (b) Matching results using BRISK; (c) Matching results using SURF; (d) Matching results using HesAffNet+HardNet.

In Fig. 8, the matching results of underwater acoustic optical images without any processing are not satisfactory. The above methods could hardly get the correct matching effects, and the number of matching pairs is very small. This is mainly because the above methods are developed based on traditional optical images and are very insensitive to acoustic images, so we will mainly evaluate the results after processing by our proposed method in the next section.

Firstly, take one of these samples to specifically express the three processes of matching and visualize the detection result of feature points and feature areas.

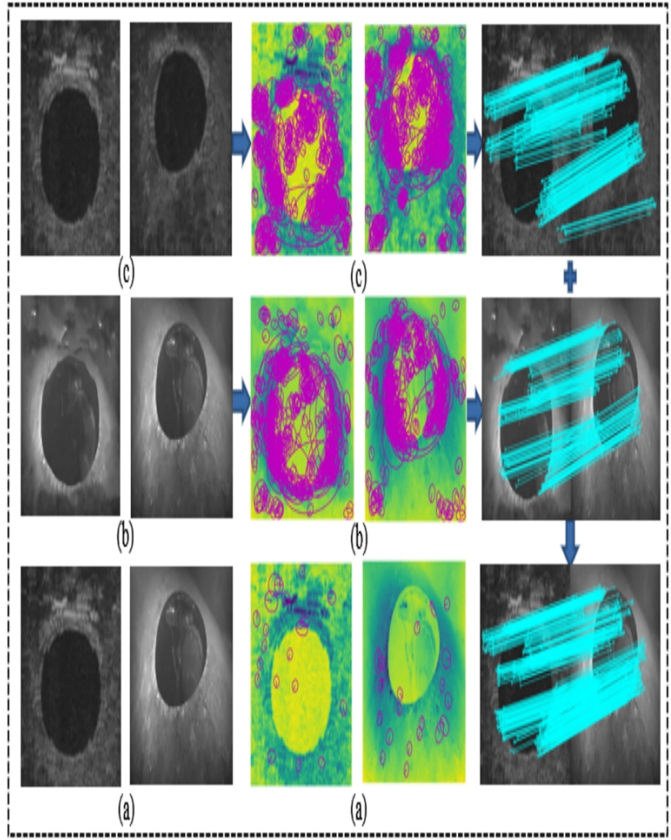

Fig. 9. Visualization of the effect of attribute transfer in the image matching process, from left to right, there are image pairs display, feature point detection schematic diagram, and matching effect schematic diagram in sequence: (a) Original image pair; (b) Image pair with optical style; (c) Image pair with acoustic style.

It can be seen from Fig. 9 that due to the huge difference in imaging principles between acoustic sensors and optical sensors, it is difficult to detect feature points on the original image. When the image undergoes attribute transfer, the imaging difference of the image is almost eliminated, and very dense feature points can be detected. In addition, because of the high quality of optical imaging. It also could be seen that the matching accuracy of image pair in optical style is higher than that in acoustic style. We fused the matching pair of the acoustic image pair and the optical image pair, and mapped them to the original image after the error elimination operation.

Each data sample in the experiment is tested according to the method shown in Fig. 9. The default image attribute transfer operation has completed, using the algorithm of RANSAC [51] and the cross-check during matching process, *ratio* is set at 0.8. The matching results as shown in Fig. 10 and the detailed evaluation results from image pair 1 to image pair 6 are presented in Table 1, Table 2 and Table 3.

TABLE I
RESULTS OF EVALUATION

| Image | Image pair 1 | | | | Image pair 2 | | | |
|---|---|---|---|---|---|---|---|---|
| Eval Meth | GM | INL | MA | RT (s) | GM | INL | MA | RT (s) |
| Tran+ SIFT | 67 | 36 | 0.54 | 0.13 | 68 | 51 | **0.75** | **0.12** |
| Tran+ BRISK | 2 | 1 | 0.50 | **0.12** | 13 | 7 | 0.54 | 0.14 |
| Tran+ SURF | 221 | 123 | 0.56 | 0.25 | 159 | 096 | 0.60 | 0.26 |
| Tran+ AffNet HardNet | **842** | **487** | 0.58 | 2.83 | **713** | **413** | 0.58 | 2.77 |

Eval = Evaluation, Meth = Method; Tran = after attribute transfer

TABLE 2
RESULTS OF EVALUATION

| Image | Image pair 3 | | | | Image pair 4 | | | |
|---|---|---|---|---|---|---|---|---|
| Eval Meth | GM | INL | MA | RT (s) | GM | INL | MA | RT (s) |
| Tran+ SIFT | 233 | 145 | 0.62 | 0.36 | 102 | 60 | **0.59** | **0.15** |
| Tran+ BRISK | 44 | 33 | **0.75** | **0.18** | 12 | 7 | 0.58 | 0.16 |
| Tran+ SURF | 191 | 097 | 0.51 | 0.28 | 189 | 87 | 0.46 | 0.28 |
| Tran+ AffNet HardNet | **422** | **286** | 0.68 | 2.71 | **396** | **227** | 0.57 | 2.54 |

Eval = Evaluation, Meth = Method; Tran = after attribute transfer

TABLE 3
RESULTS OF EVALUATION

| Image | Image pair 5 | | | | Image pair 6 | | | |
|---|---|---|---|---|---|---|---|---|
| Eval Meth | GM | INL | MA | RT (s) | GM | INL | MA | RT (s) |
| Tran+ SIFT | 243 | 243 | **1.00** | 0.12 | 136 | 130 | **0.96** | 0.22 |
| Tran+ BRISK | 169 | 168 | 0.99 | **0.11** | 33 | 31 | 0.94 | **0.16** |
| Tran+ SURF | 1097 | 1096 | 0.99 | 0.25 | 214 | 185 | 0.86 | 0.25 |
| Tran+ AffNet HardNet | **4388** | **4388** | 1.00 | 2.69 | **526** | **493** | 0.94 | 2.50 |

Eval = Evaluation, Meth = Method; Tran = after attribute transfer

In image pair 1 to image pair 4, they prove that when the underwater images have affine transformations, the matching results obtained by our method have the highest quality. In image pair 5 and image pair 6, when there are only imaging differences, our method can also get the densest matching pairs.

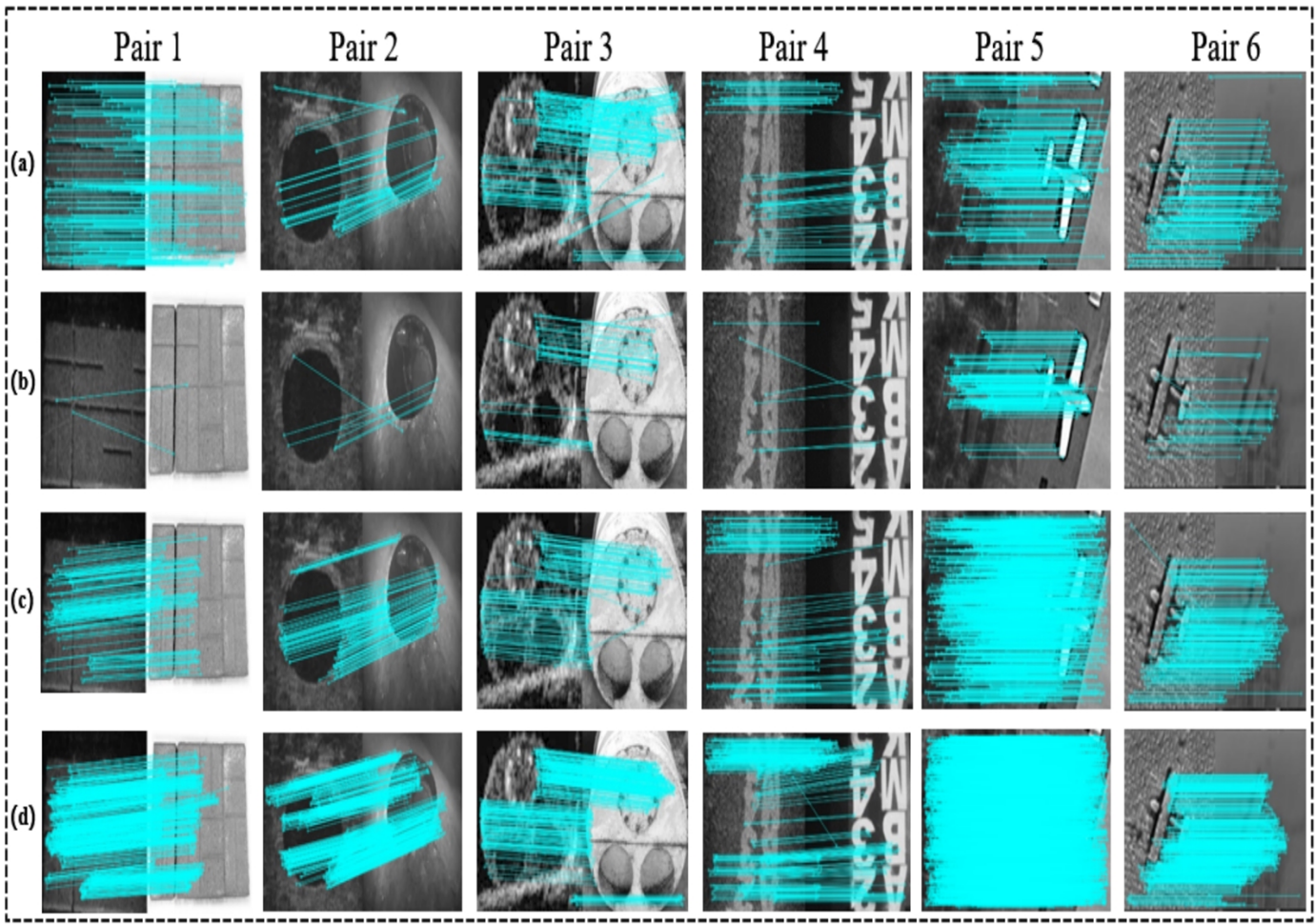

Fig. 10. This is a schematic diagram of the matching results of the images after attribute transfer: (a) Matching results using SIFT; (b) Matching results using BRISK; (c) Matching results using SURF; (d) Matching results using HesAffNet+HardNet.

## VI. DISCUSSION

For the underwater acousto-optic image matching task, we have solved it from the following aspects: (1) use the image attribute transfer method to maximize the elimination of the difference between the underwater acoustic image and optical image. (2) introduce the advanced HardNet descriptor, and use the method based on local features to match the acoustic and optical images. Experimental verification: although HardNet has not been deliberately trained on underwater sonar and optical datasets, it still shows the impressive results, and could overcome the viewing angle and background changes caused by the sonar detection process. The proposed method does not make any assumptions about the type of input acousto-optic image, nor does it require manual feature intervention and data preprocessing. The matching method is close to end-to-end.

This letter utilizes real underwater acousto-optic data and simulation data for testing. The experimental results show that our method has excellent effects and strong robustness. In the process of attribute transfer, due to the poor imaging effects of sonar, chaos will inevitably appear in the edge area of the image structure. Next, the algorithm of detail segmentation will be used to optimize the attribute transfer process of the sonar image to obtain a high-quality target image.

## VII. CONCLUSION

A method of applying visual attribute transfer algorithm and the learned descriptor is proposed to achieve underwater acousto-optic image matching. The algorithm could be applied to a variety of underwater operation scenarios and there is no need to superimpose complex sonar image processing methods. The experiment proves our proposed could effectively solve the matching problem of underwater acousto-optic images with high accuracy and robustness. Additionally, the method of style transfer is introduced to solve the problem of the scarcity of underwater image samples.

In the future, we will focus on further expanding the number of underwater samples and training local feature descriptors in the acousto-optic field to achieve better performances. And to further optimize the preprocessing algorithm for image attribute transfer to make it lightweight, in order to better meet the real-time requirements of underwater engineering operations and enhance the autonomy of deep submersibles.

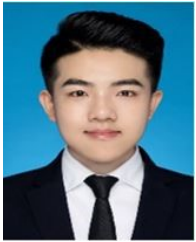

**Xiaoteng Zhou** is currently pursuing his Master's degree with the School of Ocean Engineering, Harbin Institute of Technology, China. His research interests are sonar image processing and analysis, pattern recognition, machine learning and their applications in underwater detection equipment.

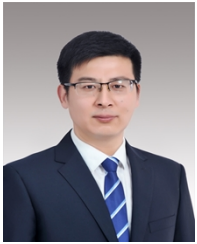

**Changli Yu** received the B.Eng. and M.S. degrees in Naval Architecture and Ocean Engineering from Harbin Engineering University, China, and the Ph.D. degree in Marine Engineering from the University of Ulsan, Korea, in 2014. He is an Associate Professor in the School of Ocean Engineering, Harbin Institute of Technology, China. His research interests include assessment of ultimate strength and safety of ships and marine structures, underwater technology and equipment, standardized vessel types for fishing vessels, and noise assessment and control of marine structures.

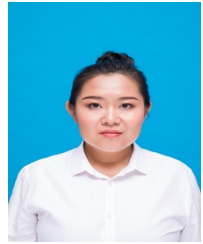

**Xin Yuan** received her Ph.D. and M.S. degree in Escuela Técnica Superior de Ingeniería y Sistemas de Telecomunicación (ETSIST) from the Universidad Politécnica de Madrid (UPM), Spain, in 2018 and 2015, separately. Her Master and Ph.D. studies are founded by the China Scholarship Council (CSC). She received her B.Eng. degree in control science and engineering from Harbin Institute of Technology (HIT), China, in 2014.

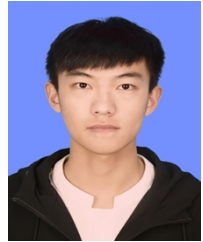

**Citong Luo** received the B.Eng. degree in Robotics Engineering from Harbin Institute of Technology, China, in 2021. He is pursuing his Master's degree with the School of Ocean Engineering, Harbin Institute of Technology, China. His research interests include the application of object detection in underwater images, and Simultaneous Localization and Mapping (SLAM).